\documentclass{article}

\usepackage[final]{neurips}

\usepackage[utf8]{inputenc} 
\usepackage[T1]{fontenc}    
\usepackage{hyperref}       
\usepackage{url}            
\usepackage{booktabs}       
\usepackage{amsfonts}       
\usepackage{nicefrac}       
\usepackage{microtype}      
\usepackage{graphicx}
\usepackage{algorithm}
\usepackage{algorithmic}
\usepackage{tabularx}
\usepackage{makecell}
\usepackage{array}
\newcolumntype{Y}{>{\centering\arraybackslash}X}
\usepackage{enumitem}

\usepackage{amsmath}
\usepackage{amssymb}
\usepackage{mathtools}
\usepackage{amsthm}

\DeclareMathOperator*{\argmax}{arg\,max}

\title{Learning, Fast and Slow: \\
A Goal-Directed Memory-Based Approach for Dynamic Environments}

\author{%
 John Chong Min Tan \\
  Department of Electrical and Computer Engineering\\
  National University of Singapore\\
  \texttt{johntancm@u.nus.edu.sg} \\
  \AND
  Mehul Motani \\
  Department of Electrical and Computer Engineering\\
  National University of Singapore\\
  \texttt{motani@nus.edu.sg} \\ 
}

\begin{document}

\maketitle

\begin{abstract}
Model-based next state prediction and state value prediction are slow to converge. To address these challenges, we do the following: i) Instead of a neural network, we do model-based planning using a parallel memory retrieval system (which we term the \textit{slow} mechanism); ii) Instead of learning state values, we guide the agent's actions using goal-directed exploration, by using a neural network to choose the next action given the current state and the goal state (which we term the \textit{fast} mechanism). The goal-directed exploration is trained online using hippocampal replay of visited states and future imagined states every single time step, leading to fast and efficient training. Empirical studies show that our proposed method has a 92\% solve rate across 100 episodes in a dynamically changing grid world, significantly outperforming state-of-the-art actor critic mechanisms such as PPO (54\%), TRPO (50\%) and A2C (24\%). Ablation studies demonstrate that both mechanisms are crucial. We posit that the future of Reinforcement Learning (RL) will be to model goals and sub-goals for various tasks, and plan it out in a goal-directed memory-based approach.
\end{abstract}

\section{Introduction}

Humans learn quickly, while Reinforcement Learning (RL) takes millions of time steps to learn how to perform tasks such as locomotion \citep{schulman2017proximal} or Atari games \citep{mnih2013playing,hafner2020mastering}. We posit that the traditional focus of maximizing reward in an optimization fashion \citep{sutton2018reinforcement} for RL would entail the need to constantly explore the environment even after solving in order to find the optimal path, leading to slow convergence to the solution path. This constant exploration may be required for optimization-based games such as chess and Go in order to continually improve \citep{silver2016mastering,silver2017mastering,schrittwieser2020mastering}, and indeed, human masters in these games spend years to perfect and hone their skill. However, in most real-life tasks such as navigation, locomotion or even deciding what to eat for lunch, optimality may not be required. Rather, fast learning and decision making should be prioritized in order to survive in a fast-paced world. Such a satisficing agent could perhaps be used in self-driving cars whereby the environment changes frequently. In such environments, a pursuit of optimality is not just sample intensive and impractical, but can be detrimental to adaptive learning as a once-optimal policy might need to be unlearned to do well should the environment change.

\begin{figure}
\begin{center}
\includegraphics[width = 0.7\textwidth]{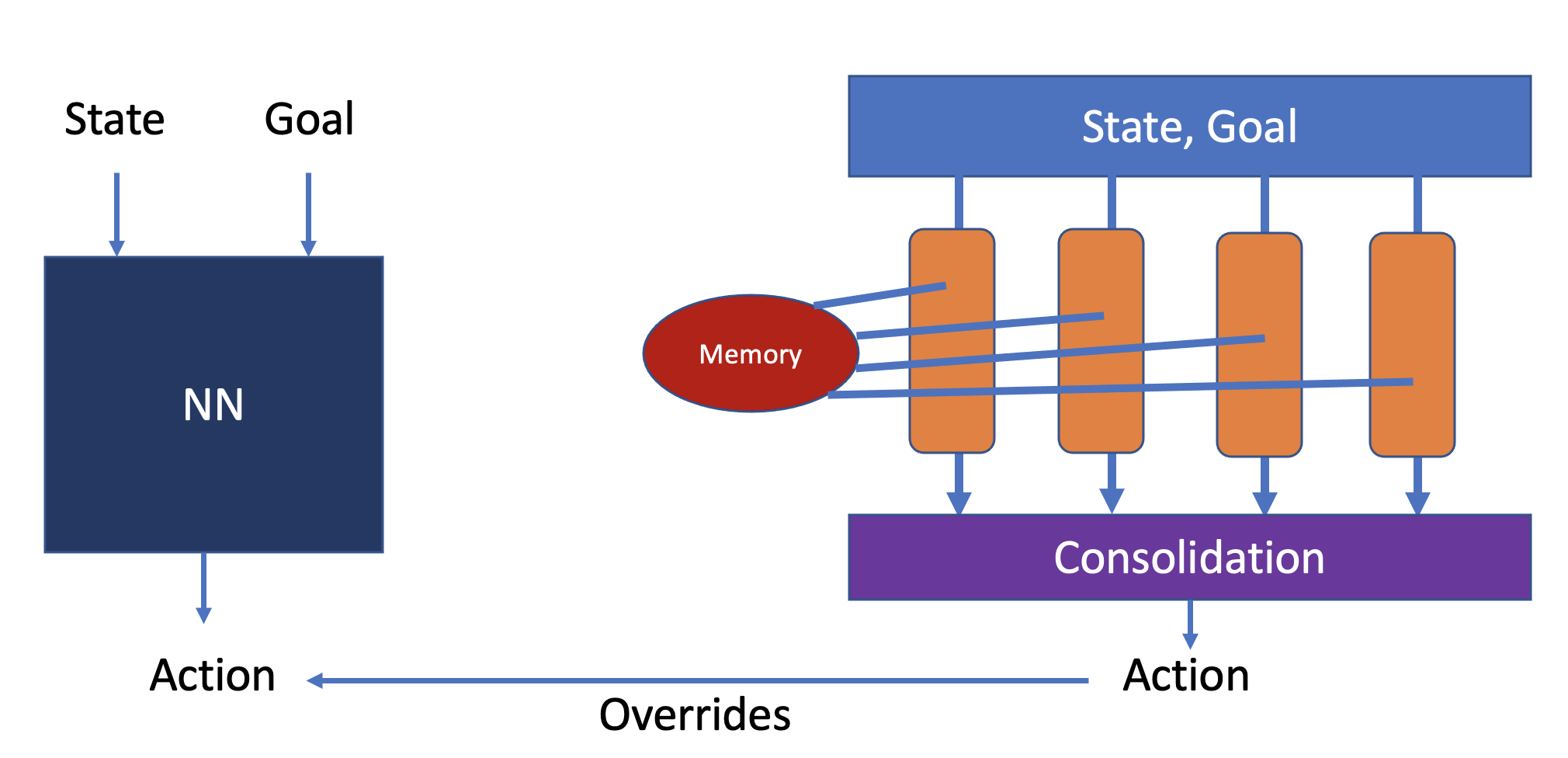}
\end{center}
\caption{Learning, Fast and Slow. \textbf{(Left)} Fast mechanism for inference using Neural Network. \textbf{(Right)} Slow mechanism for inference using parallel memory retrieval.}
\label{fig: RL fast and slow}
\end{figure}

We introduce a type of online RL which does not seek to optimize, but rather, to satisfice. When we remove optimality as a hard constraint, we can develop agents which learn and adapt faster to changing environments. Our proposed approach consists of two parts (see \textbf{Fig. \ref{fig: RL fast and slow}}):

\textbf{Goal-Directed Mechanism (Fast).} Humans are typically goal-directed, and imbuing this pursuit of a goal to an AI system could lead to efficient exploration of an environment. This is implemented via a goal-conditioned neural network.

\textbf{Memory-based Mechanism (Slow).} Humans typically use memory to guide selection of actions, and doing so can lead to finding a solution path based on past experiences. This is implemented via hash table storage and retrieval.

\section{Preliminaries - Modeling the World}

There have been a series of works that utilize world models to do next state prediction. Such model-based methods have been used successfully in MuZero for Atari games, chess, Go, shogi \citep{schrittwieser2020mastering}, as well as SimPLE \citep{kaiser2019model}, Dreamer \citep{hafner2019dream}, Dreamer v2 for Atari games \citep{hafner2020mastering} and Dreamer v3 for multiple domains \citep{hafner2023mastering}. These model-based methods are generally more sample efficient \citep{sutton2018reinforcement}, but the downside is that the world models take a long time to learn. This is notably so in MuZero which takes 80 GPU days to achieve superhuman performance in Atari games (see Table 3 in \cite{hafner2020mastering}), while a human just needs 2 hours to be able to perform sufficiently well in the games \citep{mnih2013playing}. Moreover, such a next state prediction can be very lossy, as can be seen in Fig. 5 of \cite{hafner2023mastering} where the world model prediction deviates from the ground truth after just 5 frames. 

\subsection{Difficulty of next state prediction} 
\label{Section: Next state prediction} We perform an experiment to illustrate this point more concretely. Here, we contrast the performance of next action prediction (policy network) versus next state prediction (world model) given the current state and the goal state. The environment used was either a 10x10 grid or a 20x20 grid, with the actions from the set \{Up, Down, Left, Right, Don't Move\}. We use a two-layer Multi-Layer Perceptron (MLP) with 128 nodes each and output to a final softmax layer of 5 nodes for next action prediction, and 10/20 nodes for next state prediction. We train the model using categorical cross-entropy loss using 1000 samples (See \textbf{Appendix \ref{Appendix: Action and Next State Prediction}} for more details). The correct next action and next state corresponds to the fastest next step to be taken in order to reach the goal, preferring moves along the x-axis first rather than y-axis. At epoch 50 for actions and epoch 200 for next state prediction, we introduce a change in some predictions by changing the preference to prefer moves on y-axis first rather than x-axis. We seek to find out two things: i) how fast it takes for the model's predictions to converge to the ground truth, ii) how fast the trained model takes to adapt to a prediction change.

\begin{figure}[t]
\centering
	\begin{minipage}[t]{0.4\textwidth}
		\includegraphics[width=\textwidth]{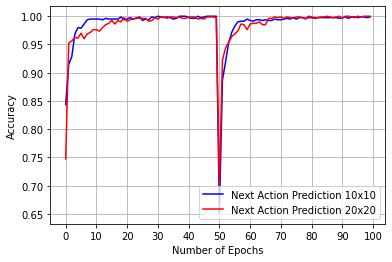}
		\caption{Accuracy of next action prediction for 10x10 grid (blue) and 20x20 grid (red) using 1000 samples}
		\label{fig:next_action_prediction}
	\end{minipage}%
	\hfill
	\begin{minipage}[t]{0.4\textwidth}
		\includegraphics[width=\textwidth]{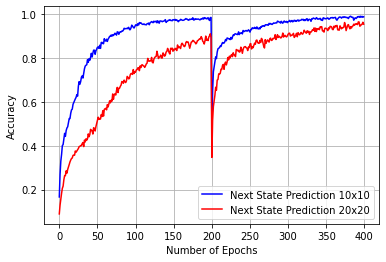}
		\caption{Accuracy of next state prediction for 10x10 grid (blue) and 20x20 grid (red) using 1000 samples}
		\label{fig:next_state_prediction}
	\end{minipage}
	\hfill
\end{figure}

\subsection{Results for next action/state prediction} 
\textbf{Figs. \ref{fig:next_action_prediction} and \ref{fig:next_state_prediction}} detail the accuracy of predicting the next action and state respectively in a 10x10 or 20x20 grid world.

\textit{Q1. How fast does the model take to converge?}

For the 10x10 grid, we can see that the next action prediction only takes 10 epochs to converge (accuracy of 0.99 and above), while the next state prediction takes approximately 200 epochs. For the 20x20 grid, the next action prediction takes about 20 epochs to converge, while the next state prediction has not converged even after 200 epochs. The next state prediction takes almost 20 times as long, just judging by the results of the 10x10 grid. This highlights the inefficiencies of trying to learn the next state prediction from observation.

\textit{Q2. How fast does the model adapt to a prediction change?}

For both the 10x10 grid and 20x20 grid, the prediction change of the next action and next state was learned in a quicker time than the time it took for convergence originally. Notably, it only took 5 and 15 epochs to converge for the actions for the 10x10 grid and 20x20 grid respectively. Correspondingly, it took 150 and more than 200 epochs for the next state prediction to converge. The next state prediction takes almost 30 times as long, just judging by the results of the 10x10 grid. This again highlights the inefficiencies of trying to learn the next state prediction from observation.

\textbf{Interpretation of Results.}
The results show that next action prediction is much faster to learn than next state prediction, and we design our RL agent with this in mind. We will want to utilize this next action prediction in the form of a goal-conditioned neural network to predict actions, very similar to the policy network in that of Actor-Critic models. Also, we will not want to use neural networks to do next state prediction, and instead, utilize memory retrieval to do model-based planning.

\section{Incorporating Goals - Reward is not enough}
Maintaining a value of each state (or state-action pair) is typical in RL and can serve as a way to cache intermediate states. If the environment is unchanging, this can be useful for determining how good the next state is, such as in unchanging board environments like Go \citep{silver2016mastering}. However, since correctly evaluating each state's value takes time, it will be difficult to evaluate the value exactly if the environment is constantly changing. Moreover, even within the same environment, a variant of the task usually entails a different reward function (i.e. navigating to different locations), and this leads to added difficulties in learning the state value function. In such situations, it may be better to specify the problem not in terms of maximizing reward, but rather, to fulfill a goal.

In contrast to the standpoint by \cite{silver2021reward} that rewards are enough, and are ``sufficient to express a wide variety of goals", we posit that rewards are not crucial to shape an agent's behavior if there is already a sufficiently good way to model goals into the system. For cases such as doing well in an Atari game with arbitrary external rewards associated with each state, we may need to model such a value function to do well. However, if we are thinking about navigation in real-life whereby we already have an end-goal in mind, such value modeling may not be necessary. 

Indeed, for sparse reward settings \citep{ecoffet2019go, ecoffet2021first}, the usefulness of reward as a signal is diminished and curiosity-based intrinsic rewards such as that in Intrinsic Curiosity Module (ICM) \citep{pathak2017curiosity} may be needed to boost the reward signal. The fact that an agent requires alternate rewards to learn in a sparse reward setting hints that reward alone is not sufficient for decision making.

It is also worth highlighting that even for cases whereby reward is successfully used to solve the problem, for instance in Atari games, sticky actions (the next action has a high chance of repeating the previous one) may still be required to explore sufficiently large parts of the environment in order to solve it \citep{ecoffet2019go,ecoffet2021first}. In fact, this sticky action is reminiscent of an agent with a goal and heading straight towards it, and is very different from traditional reward-based explore-exploit agents which tend to display erratic behavior as they may sometimes explore instead of exploit while heading towards an objective.

Hence, we posit that in order to have efficient learning for RL, it is necessary to include some form of goal-directed behavior. In fact, numerous works have utilized a form of goal-based learning \citep{schaul2015prioritized, andrychowicz2017hindsight, warde2018unsupervised, colas2019curious}. Here, we propose to do this using a goal-conditioned neural network to predict the next action. 

\section{Memory for efficient learning}
Traditional RL systems just keep track of scalar rewards. This is typical in TD-Learning or Q-Learning \citep{sutton2018reinforcement} or their neural net equivalents such as Deep Q Networks \citep{mnih2013playing} or Proximal Policy Optimization (PPO) \citep{schulman2017proximal}. Recently, systems which leverage external memory, such as Go-Explore or its variants \citep{ecoffet2019go,ecoffet2021first,minusing}, has been shown to lead to improvements over just traditional reward-based mechanisms for Reinforcement Learning. In Go-Explore, the memory stores the trajectory of the shortest path and best reward accumulated so far for any visited state. Utilizing such a memory can lead to faster identification of promising states than just relying on a value estimate alone. We need not follow the exact memory mechanism deployed in these works, but just incorporate the idea of leveraging external memory for more efficient learning than just using the neural network weights.

Combining external memory with cognitive architectures has also been done in work such as Soar \citep{laird2019soar}. A memory retrieval mechanism based on state similarity to infer value is also done in \cite{botvinick2019reinforcement}. More recently, there has been work which uses large scale memory retrieval for learning in Go, which can achieve better win rates by just changing the external dataset without even changing the parameters of the agent \citep{humphreys2022large}. We seek to build upon this work and instead of just treating external memory as a static database, we add and remove memories according to the agent's experience in order to make the agent more performant and adaptable to a changing environment.

\subsection{Memory as a proxy to world models}
If we do not need to pursue optimality, we can leverage external memory for world modeling instead of learning the exact transition probabilities between states. Using external memory for world modeling has a few key advantages: 

\begin{enumerate}[leftmargin=*, nosep, wide=0pt]

\item It solves the intractability problem of probability distributions if there are unbounded number of outcomes, as probability can just be calculated on the small subset of transitions within the memory

\item It is quick to update and a change in the stored memory can immediately lead to a change in agent behavior

\item The memory can be dynamically adapted to be in line with the agent's environment - we do not need to model the entire Markov Decision Process (MDP); we just need to model the portion which is relevant for the agent.

\end{enumerate}

\begin{figure}[t]
\begin{center}
\includegraphics[width = 0.5\textwidth]{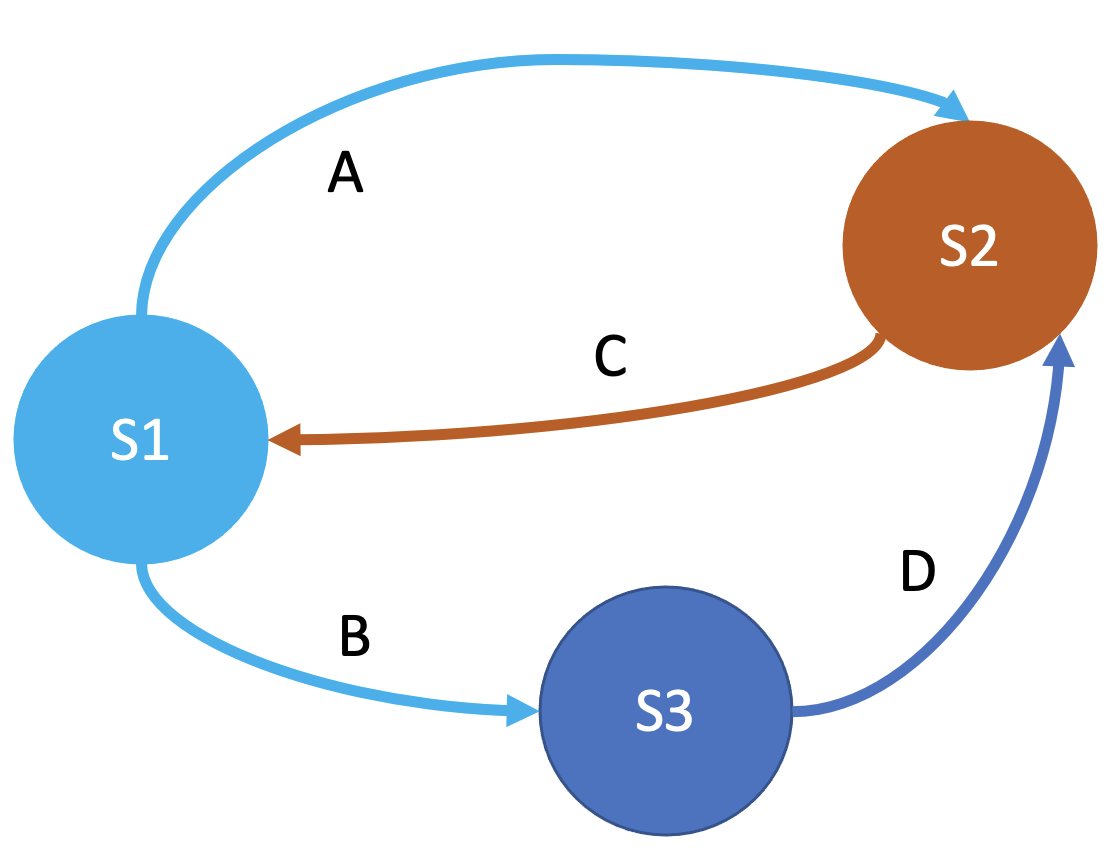}
\end{center}
\caption{A model of the world, with states labeled as S1, S2 and S3, and action transitions labeled as A, B, C, D}
\label{fig: world model}
\end{figure}

\begin{table}[t]
\caption{Memory hash table to store environmental transitions}
\label{table: MDP}
\begin{center}
\begin{tabular}{|c|c|c|} 
\hline
\textbf{Key} & \textbf{Value} & \textbf{Value}\\
\textbf{(State)} & \textbf{(Action)} & \textbf{(Next State)}\\
 \hline
1 & A & 2\\
\hline
1 & B & 3\\
\hline
2 & C & 1\\
\hline
3 & D & 2\\
\hline
\end{tabular}
\end{center}
\end{table}

Rather than modeling probabilities of the transition to next states in the MDP, we utilize a hash table with the current state as the key, and the future action and states as the values. For instance, for an MDP denoted by \textbf{Fig. \ref{fig: world model}}, the corresponding hash table is \textbf{Table \ref{table: MDP}}. 

\subsection{Different types of memory}
Typically, one refers to memory in the deep learning literature as that of the memory of the weights in the neural network, much like Long Term Potentiation in synapses of neurons \citep{lynch2004long}. However, there exists another form of memory which could be useful, and that is the kind of memory that is used in hard disks on computers - readable and writable, and provides reliable storage. While biological organisms typically use the former, the latter kind of memory has advantages of reliability and quick updating. The difference between memory in a neural network and memory of an external storage is illustrated in \textbf{Table \ref{table: Fast and Slow Mechanism}}. Neural networks and external memory retrieval/storage have their own advantages and disadvantages, and we posit that a combination of both of them is best.

\begin{table}[t]
\caption{Comparison between using neural network weights and external memory storage/retrieval}
\label{table: Fast and Slow Mechanism}
\begin{center}
\resizebox{0.7\columnwidth}{!}{
\begin{tabularx}{0.6\textwidth}{|c|X|X|} 
\hline
 \textbf{} & \textbf{Neural Network} & \textbf{External Memory Storage / Retrieval}\\
 \hline
Inference & Fast (one pass) & Slow (requires multiple lookahead retrievals)\\
\hline
Learning & Slow (requires many gradient updates) & Fast (instantaneous change by changing memory bank) \\
\hline
Generalization & Can interpolate well & Need the right abstraction space to store memory to generalize\\
\hline
Storage & Unreliable. Previously learned input-output relations may be changed with update of weights & Reliable. Previously stored memory will never be changed unless intentionally discarded\\
\hline

\end{tabularx}
}
\end{center}
\end{table}

\section{Algorithm}

Having established the benefits of both the fast goal-directed mechanism and slow memory-based retrieval mechanism, we detail a workable algorithm to implement both mechanisms in a single agent. Of note, fast and slow mechanisms have been analyzed for various domains \citep{daniel2017thinking, anthony2017thinking, botvinick2019reinforcement, pham2021dualnet}, but ours is unique for the case of online RL.

We begin with an empty episodic memory and overall memory bank. At the beginning of each episode, we reset the episodic memory bank, while allowing the overall memory bank to carry over from previous episodes.

\textbf{Goal-Directed Exploration.} Firstly, our agent needs to determine an action to take given the current state and the goal state. One way to do this will be to choose the action directly from the goal-directed neural network. This network will take in a start state and goal state as inputs, and output the probabilities of taking the next action via a softmax layer output over all the discrete actions. Our model uses 2 MLP layers of 128 nodes as the hidden layers. Mathematically, $p = f(s|g, \theta)$, where $p$ is the probability vector, $f(\cdot)$ is a learnable function mapper parameterized by the neural network weights $\theta$, $s$ is the start state, and $g$ is the goal state. We treat these probabilities as the exploitation value, and add in count-based exploration similar to that in Upper Confidence Bounds for Trees (UCT) in Monte Carlo Tree Search (MCTS) \citep{browne2012survey}. The next action will then be given by:
\begin{equation}
\label{eqn: explore-exploit}
a^* = \argmax_a (p(a) - \alpha \sqrt{numvisits(a)}),
\end{equation}
where $p(a)$ is the probability of each action generated by the goal-directed network, $\alpha$ is the exploration constant set to 1, $numvisits(a)$ is the number of times the action $a$ has been sampled from the current state $s$ and is retrieved from episodic memory.

The beauty of this mechanism is that the goal-directed neural network can serve as a compass to guide the initial action. The action may not be the best possible one, but it just needs to be approximate, much like finding how to get to a tower in the middle of a forest and just making a first step towards the tower based on its general direction. Initially, we purely use the goal-directed mechanism as a guide, as the exploration term will be 0 when there is no memory of the current state in the episodic memory. Should the sequence of actions not achieve the desired results and we return to one of the already explored states in the episodic memory, it can then be influenced by the exploration term as it will bias actions that are not tried before.

\textbf{Memory-based Retrieval and Planning.} Secondly, we will query the memory-based retrieval of a sequence of actions to see if we are able to reach the goal state. This memory-based retrieval is done in parallel across $B$ multiple branches, much like how parallel processing is done in minicolumns of the neocortex \citep{edelman1982mindful}. Each branch will match the current state to memory and retrieve the corresponding next state and action. They will continue to match until maximum lookahead depth $D$ is reached or until the goal state is found. We then select the branch with the shortest trajectory to the goal state, if there is a found trajectory. The algorithm for memory retrieval is detailed in \textbf{Algorithm \ref{Alg: Memory Retrieval}}. If we manage to find a trajectory to the goal state, we then take the first action of this trajectory and override the action found by the goal-directed mechanism, as this action is found by lookahead and hence more precise. Note that we intentionally only use memory to obtain the next state for lookahead and not a neural network next state predictor. This is because such a next state predictor takes a long time to converge and using it for planning may lead to lossy lookahead, as explored in \textbf{Section \ref{Section: Next state prediction}}.

\textbf{Perform Action.} Next, we perform the desired action and obtain the next state and reward from the environment. 

\textbf{Updating Memory Bank.} We update the episodic memory and the overall memory with this transition. The key of the memory transition is the current state, and the values are the action and the next state, as shown in \textbf{Table \ref{table: MDP}}. In order to cater for changes in dynamics of the environment, we remove all stored memories in the episodic memory and overall memory that conflict with the current transition (e.g. if a State 1 and Action 1 currently leads to State 2, we remove all memories with State 1 and Action 1 not leading to State 2). This also has the added effect of increasing the exploration bias in \eqref{eqn: explore-exploit} for wrongly predicted states and hence could serve a similar function as ICM.

\begin{figure}[t]
\begin{center}
\includegraphics[width = 0.8\textwidth]{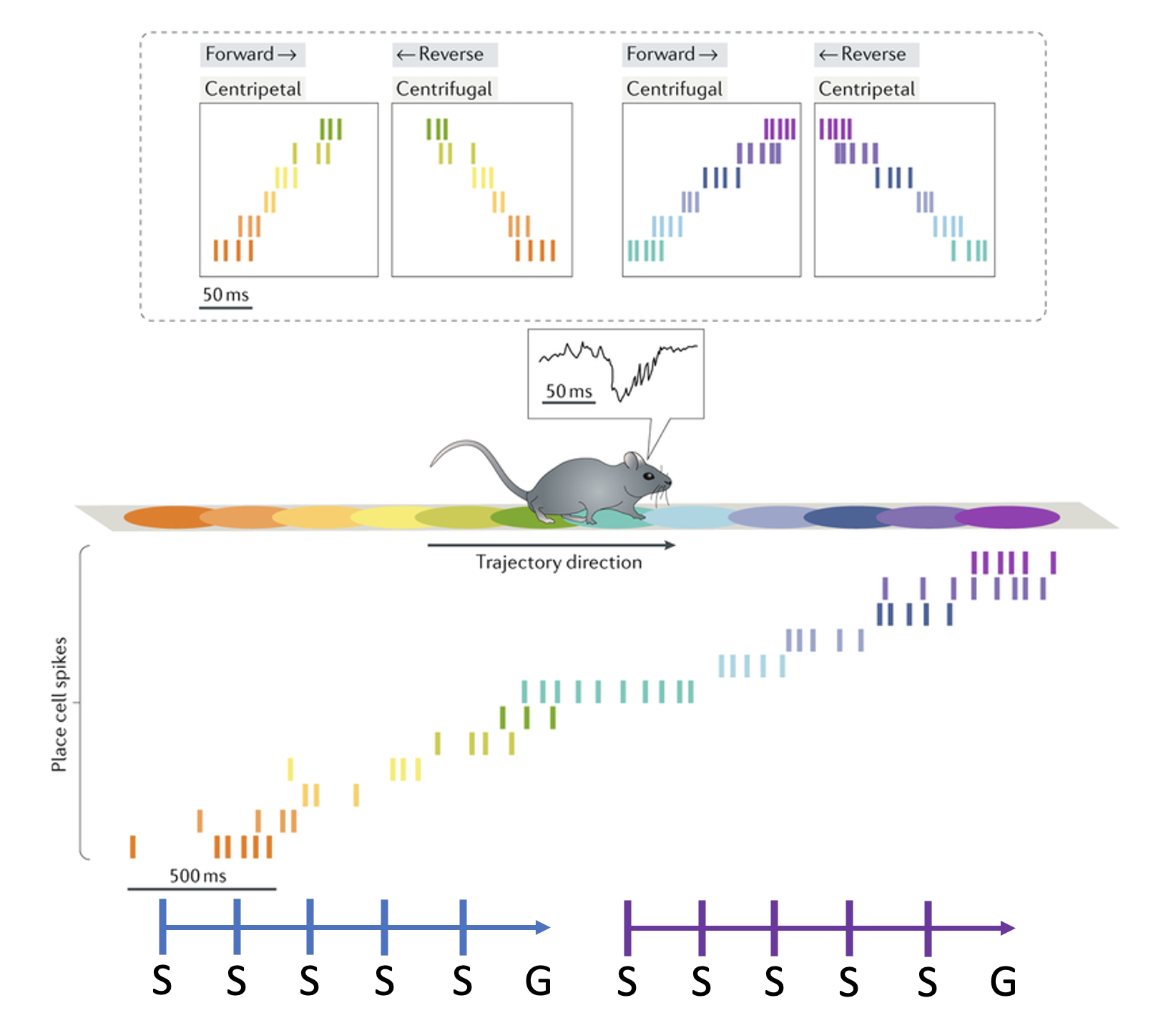}
\end{center}
\caption{Hippocampal replay in mice, which showcases forward play (pre-play) and reverse play (replay), which are involved in memory retrieval and consolidation for processes such as decision-making. Extracted from Fig. 2 of \citep{joo2018hippocampal}, with additional illustrations of a blue and purple line for goal-directed learning at the bottom ($S$ denoting start state, and $G$ denoting goal state). There is replay occurring for both 1) past visited states and 2) future imagined states. We use these insights in designing \textbf{Algorithm \ref{Alg: Hippocampal Replay}} for consolidating learned experiences. We utilize this replay to learn a goal-directed policy 1) with any state along the past trajectory as the start state and the goal state as the current state (blue line), and 2) with any state along the future imagined trajectory (if any) as the start state and the goal state as the actual goal state (purple line).}
\label{fig: hippocampal replay}
\end{figure}

\textbf{Updating Goal-Directed Neural Network.} Hippocampal replay (see \textbf{Fig. \ref{fig: hippocampal replay}}) has been known to help with memory consolidation and decision making \citep{joo2018hippocampal}. Previous works have attempted to model hippocampal replay by sampling from a replay buffer to learn the transitions \citep{mnih2013playing,mnih2015human, schaul2015prioritized}. For efficient learning, we posit that hippocampal replay should also be used to train the goal-directed neural network. The algorithm for hippocampal replay is detailed in \textbf{Algorithm \ref{Alg: Hippocampal Replay}}.

Overall, we can keep repeating the entire algorithm until the episode is completed (i.e. reward 1 attained), or until a certain amount of time steps are reached. The overall goal-directed memory-based algorithm is detailed in \textbf{Algorithm \ref{Alg: Fast and Slow}} in \textbf{Appendix \ref{appendix: Fast and Slow Algorithm}}. Our source code is made publicly available at \href{https://github.com/tanchongmin/Learning-Fast-and-Slow}{https://github.com/tanchongmin/Learning-Fast-and-Slow}.

\begin{algorithm}[t]
\caption{Memory Retrieval for Lookahead Planning}
\label{Alg: Memory Retrieval}
\begin{algorithmic}[1]
\REQUIRE Current state $s$, goal state $g$
\STATE Parallel process with $B$ branches (we use 100)
\STATE Each branch uses the state $s$ as query and retrieves the next state $s'$ and action $a$ from memory (if there are multiple retrieved memories, randomly select one)
\STATE Each branch repeats this process of memory retrieval with the subsequent state $s'$ as query for up to $D$ lookahead depth (we use 20) or until goal state $g$ reached
\STATE Consolidate all the trajectories which reach the goal state $g$ and pick the shortest
\STATE The action of the shortest trajectory (if any) will be used to override the action generated from the goal-directed mechanism
\end{algorithmic}
\end{algorithm}

\begin{algorithm}[t]
\caption{Hippocampal Replay}
\label{Alg: Hippocampal Replay}
\begin{algorithmic}[1]
\STATE Pre-Play: Consolidate the list of states in the trajectory. One trajectory is the past visited state trajectory from the start state to the current state, and the other trajectory (if found by memory retrieval mechanism) is the future unvisited state trajectory from current state to goal state
\STATE Replay: Use the last state of each trajectory as the goal state $g$, and every other state of the trajectory as the start state $s$ to form $(s,g)$ pairs for input, with the action $a$ taken at each state as output. This is used to train the goal-directed neural network
\end{algorithmic}
\end{algorithm}

\section{Experimental Setup}

\subsection{Considerations}
\textbf{Online Learning.} The key aim of the experiment was to evaluate the performance of an online learning agent. Hence, there is no training and testing phase and we evaluate the agent starting from the very first episode. 

\textbf{No Oracle World Model.} We want to provide the agent with minimal hints or guidance to make it realistic. As such, there is no perfect world model given to the agent for use for planning - the agent has to learn about the world from its interactions, and it has to learn it fast. 

\subsection{Environment}
\begin{figure}[t]
\begin{center}
\includegraphics[width = 0.8\textwidth]{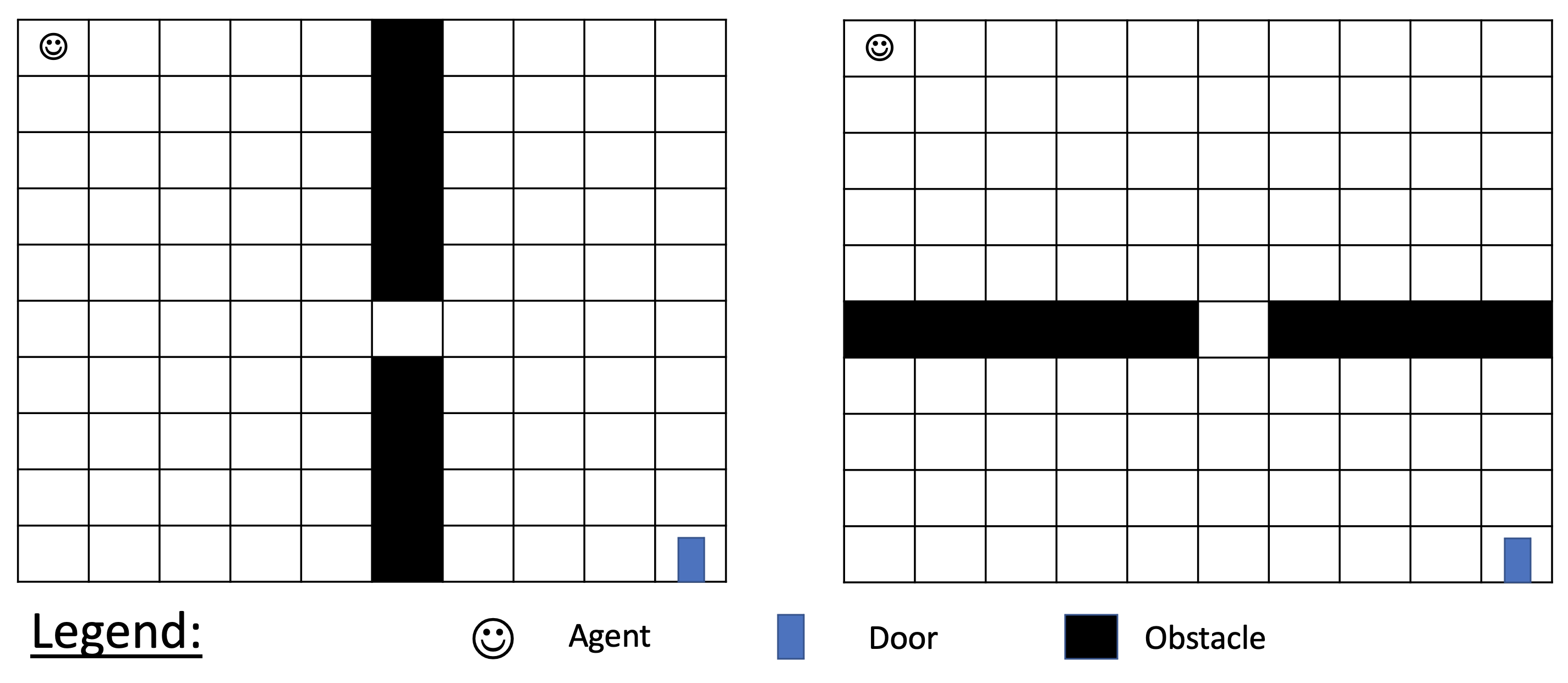}
\end{center}
\caption{A sample maze environment of size 10x10. By default, the agent's start state is at the top left and the door is at the bottom right, but it can be varied. \textbf{(Left)} Obstacles before episode 50 form a vertical wall with a gap in the center across the mid-point. \textbf{(Right)} Obstacles after episode 50 from a horizontal wall with a gap in the center across the mid-point.}
\label{fig: maze}
\end{figure}

The environment used is a 2D grid world, where there are $n$ by $n$ squares, where $n$ is the grid size. There are also some grid squares which are denoted as obstacles and are not traversable. The agent starts off at a grid square and is supposed to head towards the door (goal) position. We have two configurations of the environment used:

\begin{enumerate}[leftmargin=*, nosep, wide=0pt]

\item \textbf{Static.} There are no obstacles. The start point is at $(0, 0)$ (top left) and end point is at $(n-1, n-1)$ (bottom right). This is to evaluate learning on typical RL environments.

\item \textbf{Dynamic.} The obstacles change mid-way (episode 50), and the start and end points vary randomly with each episode. This is a difficult environment to evaluate learning on a continuously changing environment, which is not frequently studied in RL. See \textbf{Fig. \ref{fig: maze}} for an illustration. 

\end{enumerate}

\textbf{State Space.} The agent is provided with both its own position and the door (goal) position.

\textbf{Reward.} This is a sparse reward environment and the agent will only be counted as completing the episode and receive a reward of 1 if it manages to reach the door before $n\times n$ time steps. Otherwise, it will receive a reward of 0.

\textbf{Action Space.} The available action space is discrete from the set \{Up, Down, Left, Right\}. There is no wraparound, and the agent will remain in its existing position should it collide with the edges of the grid or with an obstacle. 

\begin{figure}[t]
\centering
	\begin{minipage}[t]{0.4\textwidth}
		\includegraphics[width=\textwidth]{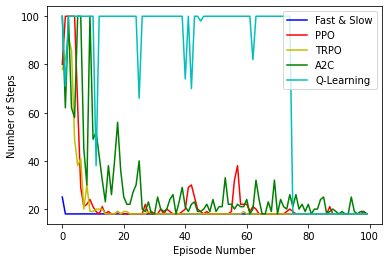}
		\caption{Steps per episode of the agents on a static 10x10 navigation task across 100 episodes}
		\label{fig:all_stars_macro}
	\end{minipage}%
	\hfill
	\begin{minipage}[t]{0.4\textwidth}
		\includegraphics[width=\textwidth]{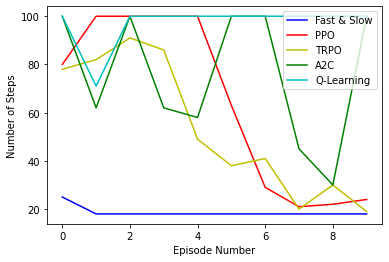}
		\caption{Steps per episode of the agents on a static 10x10 navigation task across first 10 episodes}
		\label{fig:all_stars_micro}
	\end{minipage}
	\hfill
\end{figure}

\subsection{Agents}
We use the following agents:
\begin{enumerate}[leftmargin=*, nosep, wide=0pt]

\item\textbf{Fast \& Slow Agent.} This is the proposed goal-directed (fast), memory-based (slow) agent. We use lookahead depth of 20 and 100 parallel branches for memory retrieval.

\item\textbf{Actor-Critic Agents - PPO, TRPO, A2C.} We use three competitive on-policy actor critic algorithms - PPO \citep{schulman2017proximal}, Trust Region Policy Optimization (TRPO) \citep{schulman2015trust} and Advantage Actor Critic (A2C) \citep{mnih2016asynchronous}. We use Stable Baselines 3 \citep{raffin2021stable} for reliable re-implementations of these RL algorithms. In order to give these methods the best performance in our environment, we do grid search over the following learning rates: $[0.1, 0.01, 0.001, 0.0001]$ as well as their initial default values and select the best performing one for our environment. The eventual learning rates selected were 0.0003 for PPO (default), 0.001 for TRPO (default) and 0.0001 for A2C.

\item\textbf{Q-Learning Agent.} This agent uses Q-learning, with random action selection for first few episodes, and thereafter greedy action selection. The number of episodes for random selection was selected using grid search over the entire integer interval from 0 to 100. This serves as a baseline for the efficacy of value-based methods. Note that we did not use Deep Q Network (DQN) \citep{mnih2013playing} as experiments with it failed to learn within 100 episodes, which suggests that DQN is more sample inefficient than tabular Q-learning for our environment. Refer to \textbf{Appendix \ref{Appendix: Q-Learning}} for details.

\end{enumerate}

\subsection{Evaluation Criteria}
We evaluate the agents across 100 episodes purely with online training (there is no test and training split). We use two different metrics for evaluation, as detailed below:

\begin{enumerate}[leftmargin=*, nosep, wide=0pt]

\item \textbf{Solve Rate.} This is the percentage of episodes in which the agent reaches the goal. This is a proxy for adaptability.

\item \textbf{Steps Above Minimum.} This is the number of time steps the agent takes above the minimum possible (computed using Breadth First Search). If the agent fails to complete the environment, the time step will then be the maximum time step. This is a proxy for efficiency.

\end{enumerate}

\section{Results}

The steps per episode for Fast \& Slow, PPO, TRPO, A2C and Q-Learning agents for the 10x10 static environment (minimum steps is 18) is shown in \textbf{Figs. \ref{fig:all_stars_macro} and \ref{fig:all_stars_micro}}.

The solve rate and steps above minimum for Fast \& Slow, PPO, TRPO and A2C for the 10x10 dynamic environment are detailed in \textbf{Tables \ref{table: Adaptability-10x10} and \ref{table: Efficiency-10x10}} respectively. Due to the slow learning (> 50 episodes to converge) of the Q-Learning agent on the static environment, we do not evaluate it on the dynamic environment. Refer to \textbf{Appendix \ref{Appendix: Detailed Results = Main}} for the detailed results for each episode.

\begin{table}[t]
\caption{Adaptability of methods evaluated by solve rate on a dynamic 10x10 navigation task. Higher is better (in bold).}
\label{table: Adaptability-10x10}
\begin{center}
\begin{tabular}{|c|c|c|c|} 
\hline
\textbf{Agent} & \multicolumn{3}{c|}{\textbf{Solve Rate(\%)}}\\
\cline{2-4}
&First 50 episodes & Last 50 episodes & Total\\
\hline
Fast \& Slow & \textbf{88} & \textbf{96} & \textbf{92} \\
\hline
PPO & 50 & 58 & 54\\
\hline
TRPO & 56 & 44 & 50 \\
\hline
A2C & 20 & 28 & 24 \\
\hline

\end{tabular}
\end{center}
\end{table}

\begin{table}[t]
\caption{Efficiency of methods evaluated by steps above minimum on a dynamic 10x10 navigation task. Lower is better (in bold).}
\label{table: Efficiency-10x10}
\begin{center}
\begin{tabular}{|c|c|c|c|} 
\hline
\textbf{Agent} & \multicolumn{3}{c|}{\textbf{Steps Above Minimum}}\\
\cline{2-4}
&First 50 episodes & Last 50 episodes & Total\\
\hline
Fast \& Slow & \textbf{923} & \textbf{555} & \textbf{1478} \\
\hline
PPO & 2872 & 2336 & 5208 \\
\hline
TRPO & 2669 & 3001 & 5670 \\
\hline
A2C & 4032 & 3774 & 7806 \\
\hline

\end{tabular}
\end{center}
\end{table}

\textit{Q3: How does the Fast \& Slow approach compare to traditional actor-critic/value-based approaches in a static environment?}

\textbf{Adaptability.} In terms of solve rate, we can see that Fast \& Slow and TRPO are the best (100\%), followed by PPO (96\%), A2C (95\%) and then Q-learning (32\%). In fact, Q-learning requires approximately 75 episodes before it learns via random exploration, highlighting the inefficiencies of such a value-based method. The actor-critic methods perform substantially better and solve the environment within 10 episodes. This is likely because the critic network is updated by the returns-to-go and hence learn the value of each state faster than one-step Bellman updates. For the Fast \& Slow method, the ability to combine both mechanisms give it the edge, enabling it to solve the environment the fastest.

\textbf{Efficiency.} The Fast \& Slow network has the lowest steps above minimum (7), followed by TRPO (366), PPO (576), A2C (1090) and Q-Learning (5949). The superiority of Fast \& Slow is likely due to the benefit of the slow memory mechanism finding the shortest trajectory in memory, and also being able to repeat a successful solution path.

\textit{Q4: How does the Fast \& Slow approach compare to traditional actor-critic approaches in a dynamic environment?}

\textbf{Adaptability.} In terms of solve rate, we can see that Fast \& Slow performs the best (92\%), followed by PPO (54\%), TRPO (50\%), then A2C (24\%). This highlights that traditional value-based methods can be slow to converge in the presence of varying goals in each episode. Having a goal-directed approach to infer the best action given the goal as in Fast \& Slow may be the better approach for a continually changing environment. It is also to be noted that there is learning in all algorithms except in TRPO, as even when the obstacles change, the last 50 episodes still have a higher solve rate than the first 50. This shows that an explicit memory mechanism is useful for learning, and while actor-critic approaches do have some form of memory in the weights, it is not as fast to adapt to changes.

\textbf{Efficiency.} In terms of steps above minimum, we can see that Fast \& Slow performs the best (1478), followed by PPO (5208), TRPO (5670), then A2C (7806). In fact, Fast \& Slow performs so well that it takes 4 times fewer steps above minimum than the other algorithms.

\section{Ablation Studies}
Having established the superior performance of our algorithm compared to other state-of-the-art algorithms, we conduct ablation studies to understand the components of the Fast \& Slow approach. We ablate by removing the fast and/or slow mechanisms, and change the hyperparameters of the lookahead depth (5, 10, and 50) and parallel branches (10, 50 and 200) for the memory retrieval part.

The solve rate and steps above minimum for the ablation study are detailed in \textbf{Tables \ref{table: Solve Rate-Ablation-fastslow} and \ref{table: Efficiency-Ablation-fastslow}} respectively. More detailed results can be found in \textbf{Appendix \ref{Appendix: Detailed Results - Ablation}}. Note that the baseline Fast \& Slow network uses 20 lookahead depth and 100 parallel branches.

\textit{Q5: How much do the fast and slow mechanisms contribute to performance?}

We can see that both the fast and slow mechanisms are crucial, as removing either one leads to poorer performance both in terms of adaptability and efficiency, but still comparable performance to that of the actor-critic methods analyzed in \textbf{\textbf{Tables \ref{table: Adaptability-10x10} and \ref{table: Efficiency-10x10}}}. The biggest impact comes in removing both the fast and slow mechanisms, and just relying on the count-based mechanism alone is not sufficient for performance.

The fast mechanism is actually more important than the slow one for adaptability, as the solve rate without the slow is 71\% compared to 51\% without the fast. This may be because a good initial direction from the goal-directed mechanism helps a lot more than just count-based exploration to reach the end goal. However, the slow mechanism makes up for it near the end as it is able to find the goal when it is near enough. Hence, the efficiency is similar without either the fast or the slow mechanism.

\textit{Q6: How would performance vary if we change the hyperparameters of the Fast \& Slow approach?}

In general, having more lookahead depth and parallel threads help to boost the adaptability and the efficiency of the Fast \& Slow approach. This makes intuitive sense as there are more possible (shorter) trajectories to the goal state that can be found if we search deeper and with more branches, which leads to higher solve rate and efficiency.

\begin{table}[t]
\caption{Ablation study on adaptability of Fast \& Slow agent evaluated by the solve rate of the agents on a dynamic 10x10 navigation task. Higher is better (in bold).}
\label{table: Solve Rate-Ablation-fastslow}
\begin{center}
\begin{tabular}{|c|c|c|c|} 
\hline
\textbf{Agent} & \multicolumn{3}{c|}{\textbf{Solve Rate(\%)}}\\
\cline{2-4}
&First 50 episodes & Last 50 episodes & Total\\
\Xhline{1.5pt}
Baseline & 88 & 96 & 92 \\
\Xhline{1.5pt}
No Slow & 70 & 72 & 71 \\
\hline
No Fast & 52 & 50 & 51 \\
\hline
No Fast, Slow & 26 & 18 & 22 \\
\Xhline{1.5pt}
5 lookahead depth & 82 & 96 & 89 \\
\hline
10 lookahead depth & 84 & 94 & 89 \\
\hline
50 lookahead depth & 88 & \textbf{98} & \textbf{93} \\
\Xhline{1.5pt}
10 parallel threads & 88 & 96 & 92 \\
\hline
50 parallel threads & 84 & \textbf{98} & 91 \\
\hline
200 parallel threads & \textbf{90} & 96 & \textbf{93} \\
\Xhline{1.5pt}

\end{tabular}
\end{center}

\end{table}

\begin{table}[t]
\caption{Ablation study on efficiency of Fast \& Slow agent evaluated by the steps above minimum of the agents on a dynamic 10x10 navigation task. Lower is better (in bold).}
\label{table: Efficiency-Ablation-fastslow}
\begin{center}
\begin{tabular}{|c|c|c|c|} 
\hline
\textbf{Agent} & \multicolumn{3}{c|}{\textbf{Steps Above Minimum}}\\
\cline{2-4}
&First 50 episodes & Last 50 episodes & Total\\
\Xhline{1.5pt}
Baseline & 923 & 555 & 1478 \\
\Xhline{1.5pt}
No Slow & 2493 & 2677 & 5170 \\
\hline
No Fast & 2432 & 2678 & 5110 \\
\hline
No Fast, Slow & 3894 & 4146 & 8040 \\
\Xhline{1.5pt}
5 lookahead depth & 1125 & 681 & 1806 \\
\hline
10 lookahead depth & 1100 & 905 & 2005 \\
\hline
50 lookahead depth & 898 & 488 & 1386 \\
\Xhline{1.5pt}
10 parallel threads & 1297 & 930 & 2227 \\
\hline
50 parallel threads & 1080 & 528 & 1608 \\
\hline
200 parallel threads & \textbf{791} & \textbf{445} & \textbf{1236} \\
\Xhline{1.5pt}

\end{tabular}
\end{center}
\end{table}

\section{Discussion}
Overall, it can be seen that Fast \& Slow achieves significant performance gains over state-of-the-art actor-critic models and traditional value-based methods like Q-learning in a goal-based navigation environment with a quantifiable goal state. In fact, Fast \& Slow scales up well and manages to perform well in dynamic environments of larger grid sizes like 20x20 and 40x40. For 20x20, Fast \& Slow achieves 85\% solve rate compared to best actor-critic's 18\% (4.7 fold increase in performance). For 40x40, Fast \& Slow achieves a three fold improvement, which shows the benefit of our proposed method. See \textbf{Appendix \ref{Appendix: Further Experiments}} for details.

The fast and slow mechanisms are both critical for functioning - the fast goal-directed mechanism gives an overall initial direction that aids an agent with exploring a new environment, while the slow memory retrieval mechanism gives the agent the benefit of using past experience to form a trajectory to the goal in order to guide actions.

As a plus point, due to the parallelism of the memory retrieval mechanism, Fast \& Slow has competitive runtimes to existing algorithms and is able to complete an episode on the 10x10 environment in about 2-3 seconds on a COTS CPU, making it suitable for real-world deployment.

\section{Future Work}

\textbf{Multi-Agent Learning.} The beauty of the memory mechanism is that an agent need not just learn through its own experiences, but it can internalize other agents' experiences into its memory, and have their behavior policy adjusted immediately with the incorporation of the new memory. Hence, we can have multiple agents in the same environment learning from the best performing one.

\textbf{Generic Goal Setting.} In order to utilize Fast \& Slow in domains without a quantifiable goal, one way to do so will be to use Natural Language Processing (NLP) means to vectorize a goal state via Transformer-like architectures \citep{vaswani2017attention}. This has been successfully used in SayCan \citep{brohan2022can} and it can lead to generic applications of our proposed method.

\textbf{Scaling to continuous domains.} We can map our count-based approach to continuous domains by using density models \citep{bellemare2016unifying}, or seek out approaches to abstract continuous space into discrete spaces so as to apply our algorithm to continuous state/action domains.

\textbf{Memory Forgetting.} Implementing a memory forgetting mechanism such as using the Ebbinghaus forgetting curve \citep{murre2015replication} could help to bias memories towards more recent ones that are more relevant to the environment.

\section*{Acknowledgements}
This research/project is supported by the National Research Foundation, Singapore under its AI Singapore Programme (AISG Award No: AISG2-PhD-2021-01-003[T]). 


\bibliography{neurips}
\bibliographystyle{neurips}

\newpage
\appendix
\onecolumn

\section{Experiments on Next Action and Next State Prediction}
\label{Appendix: Action and Next State Prediction}

\textbf{Setup.} We randomly generate 1000 start states and goal states (start states and goal states can be the same state) in a 10x10 or 20x20 grid world. Thereafter, we have two next predicted states, one biasing the next action/next state towards the x-axis, and one biasing the next action/next state towards the y-axis. This is because in order to reach the goal, we can either traverse via the x-axis or y-axis first. Hence, by biasing one over the other, we would still get realistic predictions while serving as a shock to the network in order to test adaptability to changing ground truths.

\textbf{Action Space.} The action space is in the set \{Up, Down, Left, Right, Don't Move\}.

\textbf{Next Action Prediction Model.} The model is an Multi-Layer Perceptron (MLP) which takes in 4 input parameters (the start and goal state coordinates). This is then fed into 2 MLP layers of 128 nodes, before reaching the final softmax layer of 5 nodes, which correspond to the output probabilities of the 5 actions.

\textbf{Next Action Prediction Model.} The model is an MLP which takes in 4 input parameters (the start and goal state coordinates). This is then fed into 2 MLP layers of 128 nodes, before splitting into two branches and each branch fed into a final softmax layer of $d$ nodes to derive the x-coordinate and y-coordinate next-state output probabilities respectively. Here, $d$ is either 10 or 20 depending on the grid size of 10 or 20.

\textbf{Loss Function.} The loss function used was the categorical cross-entropy loss, which is given by $-\sum_i^C{t_i \log(s_i)}$, where $C$ is the total number of classes, $t_i$ is the ground-truth probability of the sample for class $i$ (0 if it does not belong to class $i$, 1 if it does), and $s_i$ is the predicted class probability for class $i$ from the neural network model. As such, to minimize the categorical cross-entropy loss, the model will seek to output a probability of 1 for the ground-truth class and 0 for the rest. This is a common loss function used for training classification models, which is the use case intended here as we seek to predict by classifying the next action/state.

\textbf{Training.} We use the Adam optimizer \citep{kingma2014adam} to train our neural network. We train it for 50 epochs for next action prediction and 200 epochs for next state prediction so as to let it converge. Thereafter, we introduce the change of next action/state predictions by changing the bias of the next action/state from x-axis to y-axis. Thereafter, we let it train for another 50 epochs for next action prediction and 200 epochs for next state prediction to see how well it adapts to the prediction change.

\textbf{Accuracy.} We record the model's accuracy after each epoch of training. The next state prediction accuracy is the average of the x-axis prediction accuracy and the y-axis prediction accuracy.

\newpage
\section{Fast and Slow Algorithm}
\label{appendix: Fast and Slow Algorithm}
\textbf{Algorithm \ref{Alg: Fast and Slow}} below details the overall procedure of the Fast and Slow algorithm.

\begin{algorithm}
\caption{Fast and Slow}
\label{Alg: Fast and Slow}
\begin{algorithmic}[1]
\REQUIRE $episodic\_mem$ = dict(), $overall\_mem$ = dict()
\FOR{$episode \gets 1$ to $i$}
\STATE $episodic\_mem$ = dict()
\FOR{$step \gets 1$ to $N$}
\STATE $past\_trajectory$ = []
\STATE Retrieve current state $s$ and goal state $g$ from environment
\STATE Use goal-directed network to get output action probabilities $p = f(s|g, \theta)$ given current state $s$ and goal state $g$
\STATE Use $episodic\_mem$ to get $numvisits$, the action counts from current state $s$ 
\STATE Select action based on explore-exploit equation $a^* = \argmax_a (p(a) - \alpha \sqrt{numvisits(a)})$
\STATE $future\_trajectories$ = []
\FOR{each parallel branch $B$}
\STATE $state$ = $s$
\STATE $branch\_trajectory$ = []
\FOR{$depth \gets 1$ to $D$}
\STATE Query $overall\_mem$ with key $state$ and retrieve the values $(next\_state, action)$
\STATE Randomly select one $(next\_state, action)$ tuple
\STATE $state$ = $next\_state$
\STATE $branch\_trajectory$.append(($state$, $action$))
\IF{$state$ == $g$}
\STATE $future\_trajectories$.append($branch\_trajectory$)
\ENDIF
\ENDFOR
\ENDFOR
\STATE $future\_trajectory$ = shortest trajectory in $future\_trajectories$ if there are any, else []
\STATE Override action $a$ with first action of $future\_trajectories$ (if non-empty). 
\STATE Perform action $a$ to obtain next state $s'$ and reward $r$ from environment
\STATE $past\_trajectory$.append($(s, a)$)

\STATE Update $episodic\_mem$ and $overall\_mem$ with current state $s$ as key and $(s', a)$ as value. Remove all conflicting memories with $s$ as the key and values different from $(s', a)$
\STATE Terminate episode if $r$ == 1
\STATE For each $past\_trajectory$ and $future\_trajectory$ (if non-empty), use the last state of the trajectory as the goal state $g''$, and every other state of the trajectory as the start state $s''$ to form $(s'',g'')$ pairs for input, with the action $a''$ taken at each state as output. Use the input-output pairs to train fast neural network using gradient descent to minimize cross-entropy loss of the output action
\ENDFOR
\ENDFOR
\end{algorithmic}
\end{algorithm}

\newpage
\section{Q-Learning Agent}
\label{Appendix: Q-Learning}

Consider a Markov Decision Process (MDP) defined by the tuple $(S, A, R, P)$, where $S$ represents the set of states, $A$ represents the set of actions, $R$ represents the reward function between transitions of states, given by the function $R(s_t, a_t, s_{t+1})$, $P$ represents the transition probability of going from one state to another, given by the function $P(s_{t+1}| s_t, a_t)$, and $t$ is the time step.

We perform an online learning for the Q-functions, where we only update the state-action values that the agent visits. For every state transition $(s_t, a_t, s_{t+1})$, we calculate the TD-error: 
\begin{equation}
\delta_t = r_t + \gamma\max_{a\in A}Q(s_{t+1},a) - Q(s_t, a_t),
\end{equation}

where $\gamma$ is the discount factor which we set at 0.99.

Thereafter, we perform the Q-learning update:
\begin{equation} Q(s_t, a_t) \leftarrow Q(s_t, a_t) + \alpha\delta_t,
\end{equation}

where $\alpha$ is the learning rate which we set as 1.

We use an $\epsilon$-greedy behavior policy, where we take a random action a fraction $\epsilon$ of the time, and the greedy action $a^* = \max_{a\in A}Q(s_{t+1},a)$ a fraction $1 - \epsilon$ of the time. 

For our agent, we firstly use $\epsilon=1$ for the first few episodes to explore the state space, and thereafter use $\epsilon=0$ to greedily select actions.

\newpage
\section{Detailed Results - Main}
\label{Appendix: Detailed Results = Main}

This appendix chapter details the results which give more detailed insight into how performant the various agents are in the static and dynamic 10x10 environment.

\subsection {Static 10x10 environment}

The solve rate and steps above minimum of Fast \& Slow, PPO, TRPO, A2C and Q-Learning agents for the static 10x10 navigation task are detailed in \textbf{Table \ref{table: Adaptability Efficiency-10x10}}. We can see that Fast \& Slow and TRPO perform the best for solve rate, while Fast \& Slow is significantly better than all the other agents for steps above minimum. This highlights the superior learning abilities of the Fast \& Slow agent.

\begin{table}[h]
\caption{Adaptability and efficiency of methods evaluated by the solve rate and steps above minimum respectively of the agents on a static 10x10 navigation task. Higher is better for solve rate and lower is better for steps above minimum (in bold).}
\label{table: Adaptability Efficiency-10x10}
\begin{center}
\begin{tabular}{|c|c|c|} 
\hline
\textbf{Agent} & \textbf{Solve Rate(\%)} & \textbf{Steps above minimum} \\
\hline
Fast \& Slow & \textbf{100} & \textbf{7} \\
\hline
PPO & 96 & 576\\
\hline
TRPO & \textbf{100} & 366\\
\hline
A2C & 95 & 1090 \\
\hline
Q-Learning & 32 & 5949\\
\hline

\end{tabular}
\end{center}
\end{table}

\subsection {Dynamic 10x10 environment}

The steps per episode for Fast \& Slow, PPO, TRPO and A2C agents for the dynamic 10x10 navigation task is detailed in Figs. \ref{fig:fast and slow}, \ref{fig:no slow}, \ref{fig:no fast} and \ref{fig:no fast and slow} respectively. Note that the blue line is the minimum possible steps for each episode and the green line is the actual steps taken for each episode. Here, we can see that Fast \& Slow solves the environment for the most number of episodes and solves it with a smaller number of steps than the other agents.

\begin{figure*}[h]
\centering
        \begin{minipage}[t]{0.22\textwidth}
		\includegraphics[width=\textwidth]{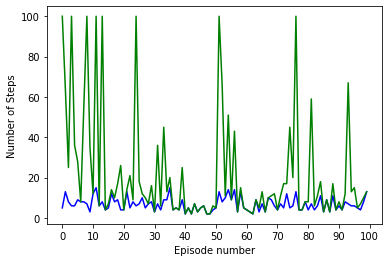}
		\caption{Steps per episode of Fast \& Slow on a dynamic 10x10 navigation task}
		\label{fig:fast and slow main}
	\end{minipage}%
	\hfill
	\begin{minipage}[t]{0.22\textwidth}
		\includegraphics[width=\textwidth]{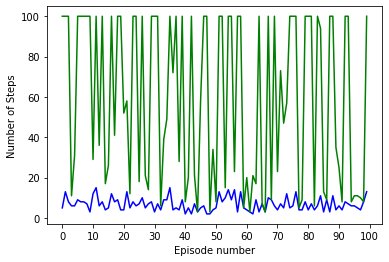}
		\caption{Steps per episode of PPO on a dynamic 10x10 navigation task}
		\label{fig:ppo}
	\end{minipage}
	\hfill
	\begin{minipage}[t]{0.22\textwidth}
		\includegraphics[width=\textwidth]{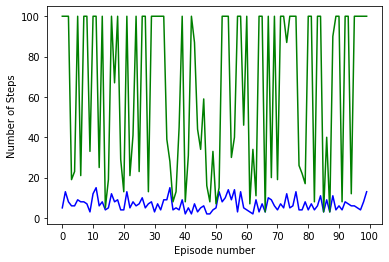}
		\caption{Steps per episode of TRPO on a dynamic 10x10 navigation task}
		\label{fig:trpo}
	\end{minipage}
	\hfill
	\begin{minipage}[t]{0.22\textwidth}
		\includegraphics[width=\textwidth]{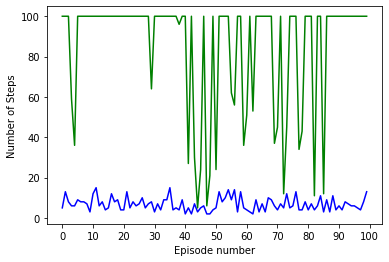}
		\caption{Steps per episode of A2C on a dynamic 10x10 navigation task}
		\label{fig:a2c}
	\end{minipage}
\end{figure*}

\newpage
\section{Detailed Results - Ablation}
\label{Appendix: Detailed Results - Ablation}

This appendix chapter details the steps per episode for various ablation studies on the Fast \& Slow agent for the dynamic 10x10 navigation task.

\subsection {Ablation: Fast and Slow components}
\begin{figure*}[h]
\centering
	\begin{minipage}[t]{0.22\textwidth}
		\includegraphics[width=\textwidth]{Pictures/fast_and_slow.png}
		\caption{Steps per episode of Fast \& Slow on a dynamic 10x10 navigation task}
		\label{fig:fast and slow}
	\end{minipage}%
	\hfill
	\begin{minipage}[t]{0.22\textwidth}
		\includegraphics[width=\textwidth]{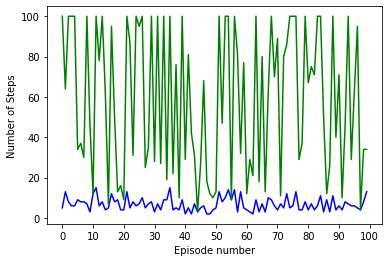}
		\caption{Steps per episode of Fast \& Slow (without slow memory retrieval mechanism) on a dynamic 10x10 navigation task}
		\label{fig:no slow}
	\end{minipage}
	\hfill
	\begin{minipage}[t]{0.22\textwidth}
		\includegraphics[width=\textwidth]{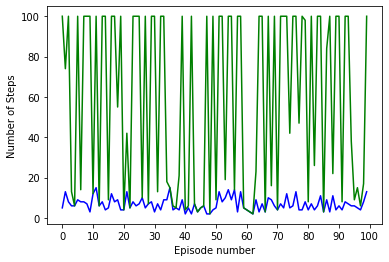}
		\caption{Steps per episode of Fast \& Slow (without fast goal-directed mechanism) on a dynamic 10x10 navigation task}
		\label{fig:no fast}
	\end{minipage}
	\hfill
	\begin{minipage}[t]{0.22\textwidth}
		\includegraphics[width=\textwidth]{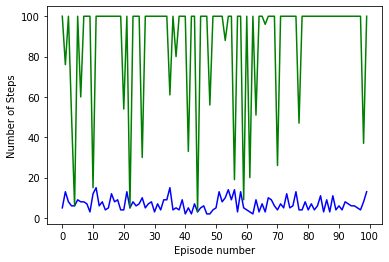}
		\caption{Steps per episode of Fast \& Slow (without both fast goal-directed and slow memory retrieval mechanisms) on a dynamic 10x10 navigation task}
		\label{fig:no fast and slow}
	\end{minipage}
\end{figure*}

The steps per episode for the baseline Fast \& Slow network, without Slow, without Fast and without both Fast \& Slow for the dynamic 10x10 navigation task are shown in Figs. \ref{fig:fast and slow}, \ref{fig:no slow}, \ref{fig:no fast} and \ref{fig:no fast and slow} respectively. Note that the blue line is the minimum possible steps for each episode and the green line is the actual steps taken for each episode. Here, we can see that both the fast and slow mechanisms are crucial for functioning, as the solve rate plummets just by removing any one of these two mechanisms.

\subsection {Ablation: Lookahead Steps}

\begin{figure*}[h]
\centering
	\begin{minipage}[t]{0.3\textwidth}
		\includegraphics[width=\textwidth]{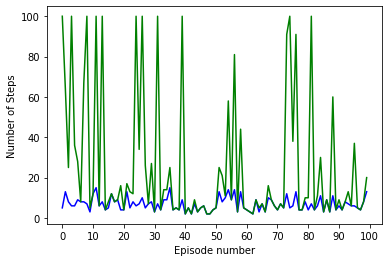}
		\caption{Steps per episode of Fast \& Slow with 5 lookahead steps on a dynamic 10x10 navigation task}
		\label{fig:lookahead_5}
	\end{minipage}
	\hfill
	\begin{minipage}[t]{0.3\textwidth}
		\includegraphics[width=\textwidth]{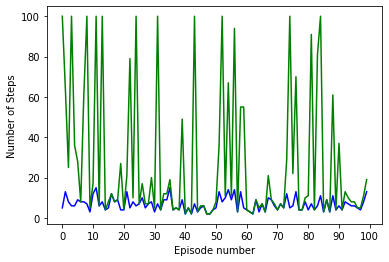}
		\caption{Steps per episode of Fast \& Slow with 10 lookahead steps on a dynamic 10x10 navigation task}
		\label{fig:lookahead_10}
	\end{minipage}
	\hfill
	\begin{minipage}[t]{0.3\textwidth}
		\includegraphics[width=\textwidth]{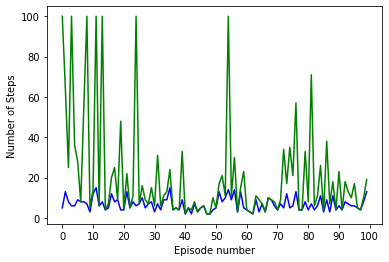}
		\caption{Steps per episode  of Fast \& Slow with 50 lookahead steps on a dynamic 10x10 navigation task}
		\label{fig:lookahead_50}
	\end{minipage}
\end{figure*}

The steps per episode for the Fast \& Slow network with 5, 10 and 50 lookahead steps for the dynamic 10x10 navigation task are shown in Figs. \ref{fig:lookahead_5}, \ref{fig:lookahead_10} and \ref{fig:lookahead_50} respectively. Note that the blue line is the minimum possible steps for each episode and the green line is the actual steps taken for each episode. Here, we can see that, in general, increasing lookahead steps leads to better performance.

\newpage
\subsection{Ablation: Parallel Branches}

\begin{figure*}[h]
\centering
	\begin{minipage}[t]{0.3\textwidth}
		\includegraphics[width=\textwidth]{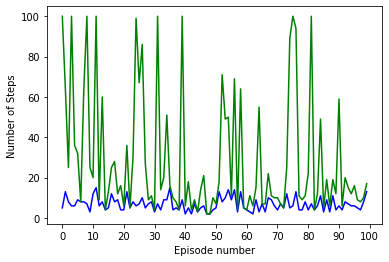}
		\caption{Steps per episode of Fast \& Slow with 10 parallel branches on a dynamic 10x10 navigation task}
		\label{fig:parallel_10}
	\end{minipage}
	\hfill
	\begin{minipage}[t]{0.3\textwidth}
		\includegraphics[width=\textwidth]{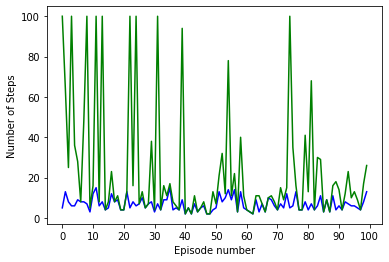}
		\caption{Steps per episode of Fast \& Slow with 50 parallel branches on a dynamic 10x10 navigation task}
		\label{fig:parallel_50}
	\end{minipage}
	\hfill
	\begin{minipage}[t]{0.3\textwidth}
		\includegraphics[width=\textwidth]{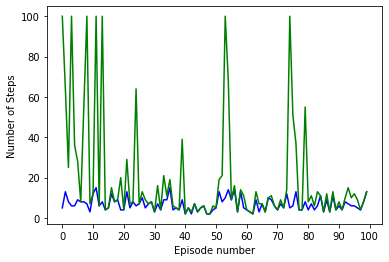}
		\caption{Steps per episode of Fast \& Slow with 200 parallel branches on a dynamic 10x10 navigation task}
		\label{fig:parallel_200}
	\end{minipage}
\end{figure*}

The steps per episode for the Fast \& Slow network with 10, 50 and 200 parallel branches for the dynamic 10x10 navigation task are shown in Figs. \ref{fig:parallel_10}, \ref{fig:parallel_50} and \ref{fig:parallel_200} respectively. Note that the blue line is the minimum possible steps for each episode and the green line is the actual steps taken for each episode. Here, we can see that, in general, increasing parallel branches leads to better performance.

\newpage
\section{Further Experiments}
\label{Appendix: Further Experiments}
Here, we detail the results of more experiments we have conducted on larger dynamic environments such as 20x20 and 40x40. This is intended as a stress test to see how scalable the various agents are.

\subsection {Dynamic 20x20 environment}

\begin{figure*}[h]
\centering
        \begin{minipage}[t]{0.22\textwidth}
		\includegraphics[width=\textwidth]{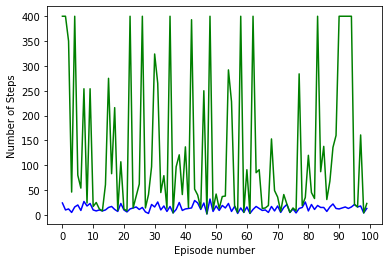}
		\caption{Steps per episode of Fast \& Slow on a dynamic 20x20 navigation task}
		\label{fig:fast and slow_20x20}
	\end{minipage}%
	\hfill
	\begin{minipage}[t]{0.22\textwidth}
		\includegraphics[width=\textwidth]{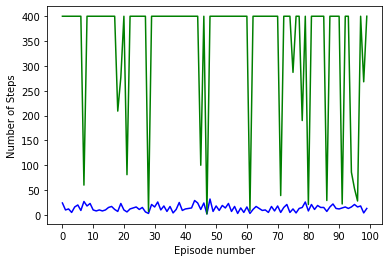}
		\caption{Steps per episode of PPO on a dynamic 20x20 navigation task}
		\label{fig:ppo_20x20}
	\end{minipage}
	\hfill
	\begin{minipage}[t]{0.22\textwidth}
		\includegraphics[width=\textwidth]{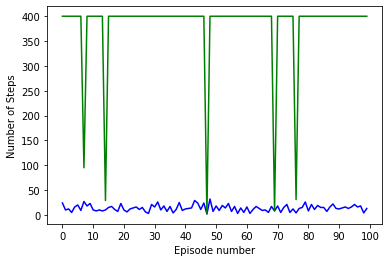}
		\caption{Steps per episode of TRPO on a dynamic 20x20 navigation task}
		\label{fig:trpo_20x20}
	\end{minipage}
	\hfill
	\begin{minipage}[t]{0.22\textwidth}
		\includegraphics[width=\textwidth]{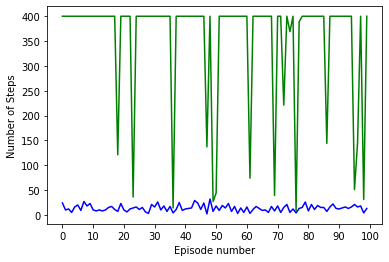}
		\caption{Steps per episode of A2C on a dynamic 20x20 navigation task}
		\label{fig:a2c_20x20}
	\end{minipage}
\end{figure*}

We seek to find out the performance of the agents in a dynamic 20x20 environment. The steps per episode for Fast \& Slow, PPO, TRPO and A2C agents for the dynamic 20x20 navigation task is detailed in Figs. \ref{fig:fast and slow_20x20}, \ref{fig:ppo_20x20}, \ref{fig:trpo_20x20} and \ref{fig:a2c_20x20} respectively. Note that the blue line is the minimum possible steps for each episode and the green line is the actual steps taken for each episode. 

The solve rate and steps above minimum for Fast \& Slow, PPO, TRPO and A2C for the 20x20 dynamic environment are detailed in \textbf{Tables \ref{table: Adaptability-20x20} and \ref{table: Efficiency-20x20}} respectively. 

Overall, we can see that the Fast \& Slow agent performs significantly better than the other agents both in terms of solve rate and steps above minimum.

\begin{table}[h]
\caption{Adaptability of methods evaluated by the solve rate of the agents on a dynamic 20x20 navigation task. Higher is better (in bold).}
\label{table: Adaptability-20x20}
\begin{center}
\begin{tabular}{|c|c|c|c|} 
\hline
\textbf{Agent} & \multicolumn{3}{c|}{\textbf{Solve Rate(\%)}}\\
\cline{2-4}
&First 50 episodes & Last 50 episodes & Total\\
\hline
Fast \& Slow & \textbf{86} & \textbf{84} & \textbf{85} \\
\hline
PPO & 14 & 22 & 18\\
\hline
TRPO & 6 & 4 & 5 \\
\hline
A2C & 10 & 22 & 16 \\
\hline

\end{tabular}
\end{center}
\end{table}

\begin{table}[h]
\caption{Efficiency of methods evaluated by the steps above minimum of the agents on a dynamic 20x20 navigation task. Lower is better (in bold).}
\label{table: Efficiency-20x20}
\begin{center}
\begin{tabular}{|c|c|c|c|} 
\hline
\textbf{Agent} & \multicolumn{3}{c|}{\textbf{Steps Above Minimum}}\\
\cline{2-4}
&First 50 episodes & Last 50 episodes & Total\\
\hline
Fast \& Slow & \textbf{6155} & \textbf{5249} & \textbf{11404} \\
\hline
PPO & 17228 & 15970 & 33198 \\
\hline
TRPO & 18218 & 18577 & 36975 \\
\hline
A2C & 17627 & 16453 & 34080 \\
\hline

\end{tabular}
\end{center}
\end{table}

\newpage
\subsection {Dynamic 40x40 environment}

\begin{figure*}[h]
\centering
        \begin{minipage}[t]{0.22\textwidth}
		\includegraphics[width=\textwidth]{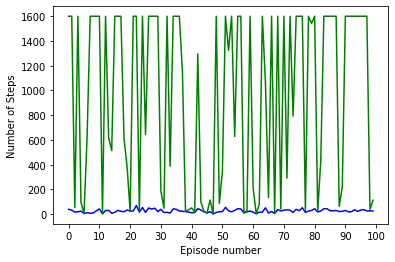}
		\caption{Steps per episode of Fast \& Slow on a dynamic 40x40 navigation task}
		\label{fig:fast and slow_40x40}
	\end{minipage}%
	\hfill
	\begin{minipage}[t]{0.22\textwidth}
		\includegraphics[width=\textwidth]{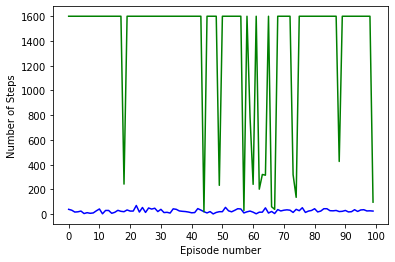}
		\caption{Steps per episode of PPO on a dynamic 40x40 navigation task}
		\label{fig:ppo_40x40}
	\end{minipage}
	\hfill
	\begin{minipage}[t]{0.22\textwidth}
		\includegraphics[width=\textwidth]{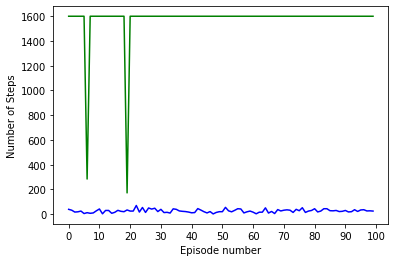}
		\caption{Steps per episode of TRPO on a dynamic 40x40 navigation task}
		\label{fig:trpo_40x40}
	\end{minipage}
	\hfill
	\begin{minipage}[t]{0.22\textwidth}
		\includegraphics[width=\textwidth]{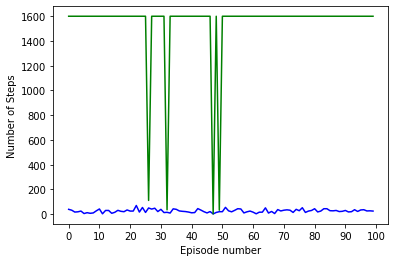}
		\caption{Steps per episode of A2C on a dynamic 40x40 navigation task}
		\label{fig:a2c_40x40}
	\end{minipage}
\end{figure*}

We seek to find out the performance of the agents in a dynamic 40x40 environment. The steps per episode for Fast \& Slow, PPO, TRPO and A2C agents for the dynamic 40x40 navigation task is detailed in Figs. \ref{fig:fast and slow_40x40}, \ref{fig:ppo_40x40}, \ref{fig:trpo_40x40} and \ref{fig:a2c_40x40} respectively. Note that the blue line is the minimum possible steps for each episode and the green line is the actual steps taken for each episode. 

The solve rate and steps above minimum for Fast \& Slow, PPO, TRPO and A2C for the 40x40 dynamic environment are detailed in \textbf{Tables \ref{table: Adaptability-40x40} and \ref{table: Efficiency-40x40}} respectively. 

Overall, we can see that the Fast \& Slow agent performs significantly better than the other agents both in terms of solve rate and steps above minimum.

\textbf{Potential improvements.} Perhaps with increasing grid size, we would need to increase memory lookahead depth and parallel threads in order to improve performance of Fast \& Slow. 

\textbf{Speed Optimization.} In order to optimize and bring down the overall training time, we train the Fast \& Slow agent's goal-directed mechanism only at the end of the episode, rather than at the end of each time step. This may bring about a slightly poorer performance at the benefit of reduced training time. As future work, we can also look into other ways to optimize the runtime, such as training the fast neural network in a separate thread and updating the model weights once every few episodes.

\begin{table}[h]
\caption{Adaptability of methods evaluated by the solve rate of the agents on a dynamic 40x40 navigation task. Higher is better (in bold).}
\label{table: Adaptability-40x40}
\begin{center}
\begin{tabular}{|c|c|c|c|} 
\hline
\textbf{Agent} & \multicolumn{3}{c|}{\textbf{Solve Rate(\%)}}\\
\cline{2-4}
&First 50 episodes & Last 50 episodes & Total\\
\hline
Fast \& Slow & \textbf{54} & \textbf{44} & \textbf{49} \\
\hline
PPO & 6 & 24 & 15\\
\hline
TRPO & 4 & 0 & 2 \\
\hline
A2C & 8 & 0 & 4 \\
\hline

\end{tabular}
\end{center}
\end{table}

\begin{table}[h]
\caption{Efficiency of methods evaluated by the steps above minimum of the agents on a dynamic 40x40 navigation task. Lower is better (in bold).}
\label{table: Efficiency-40x40}
\begin{center}
\begin{tabular}{|c|c|c|c|} 
\hline
\textbf{Agent} & \multicolumn{3}{c|}{\textbf{Steps Above Minimum}}\\
\cline{2-4}
&First 50 episodes & Last 50 episodes & Total\\
\hline
Fast \& Slow & \textbf{42707} & \textbf{50845} & \textbf{93552} \\
\hline
PPO & 74469 & 62397 & 136866 \\
\hline
TRPO & 76029 & 78628 & 154657 \\
\hline
A2C & 72553 & 78628 & 151181 \\
\hline

\end{tabular}
\end{center}
\end{table}

\end{document}